\documentclass[letterpaper, 10 pt, conference]{ieeeconf}
\IEEEoverridecommandlockouts
\usepackage{cite}
\usepackage{amsmath,amssymb,amsfonts}
\usepackage{algorithmic}
\usepackage{graphicx}
\usepackage{textcomp}
\usepackage{xcolor}
\usepackage[caption=false,font=footnotesize]{subfig}
\usepackage{booktabs}
\usepackage{hyperref}
\def\BibTeX{{\rm B\kern-.05em{\sc i\kern-.025em b}\kern-.08em
    T\kern-.1667em\lower.7ex\hbox{E}\kern-.125emX}}
    \hypersetup{
    colorlinks=true,
    linkcolor=blue,
    filecolor=magenta,      
    urlcolor=cyan,
}
\begin{document}

\title{HOPPY: An Open-source Kit for Education with Dynamic Legged Robots}

\author{Joao Ramos$^{1,2}$, 
        Yanran Ding$^{1}$,
        Young-woo Sim$^{1}$,
        Kevin Murphy$^{1}$,
        and Daniel Block$^{2}$
\thanks{Authors are with the $^{1}$Department of Mechanical Science and Engineering and the $^{2}$Department of Electrical \& Computer Engineering at the University of Illinois at Urbana-Champaign, USA. {\tt\footnotesize jlramos@illinois.edu}}}

\maketitle
\pagenumbering{gobble}

\begin{abstract}
This paper introduces HOPPY, an open-source, low-cost, robust, and modular kit for robotics education. The robot dynamically hops around a rotating gantry with a fixed base. The kit is intended to lower the entry barrier for studying dynamic robots and legged locomotion with real systems. It bridges the theoretical content of fundamental robotic courses with real dynamic robots by facilitating and guiding the software and hardware integration. This paper describes the topics which can be studied using the kit, lists its components, discusses preferred practices for implementation, presents results from experiments with the simulator and the real system, and suggests further improvements. A simple heuristic-based controller is described to achieve velocities up to $1.7m/s$, navigate small objects, and mitigate external disturbances when the robot is aided by a counterweight. HOPPY was utilized as the subject of a semester-long project for the Robot Dynamics and Control course at the University of Illinois at Urbana-Champaign. The positive feedback from the students and instructors about the hands-on activities during the course motivates us to share this kit and continue improving in the future.
\end{abstract}

\section{Introduction}

The imminent robotics revolution will employ robots as ubiquitous tools in our lives. Many machines are already being widely used in factories, assembly lines, and, more recently, in automated warehouses \cite{IEEE_Kiva}. However, most of the tasks performed by these robots are quasi-static, which means that the robot can stop mid-motion without destabilizing (falling down). In contrast, humans possess the capability to efficiently execute dynamic tasks such as running or weightlifting, where the task cannot be interrupted mid-motion. For instance, a runner cannot instantaneously freeze motion between steps without falling; an Olympic weightlifter cannot statically lift the payload above his/her head. Thus, to make robots with human-level abilities accessible in the future, we must train the next generation of roboticists to create machines that are capable of robustly performing dynamic physical tasks. Dynamic motions such as legged locomotion impose unique challenges regarding mechanical robustness, actuation saturation, control rate, state estimation, and more. However, performing experiments with dynamic robot motions is challenging because capable hardware is expensive and not readily available, and errors can quickly lead to terminal hardware damage \cite{IEEE_RobotsFalling}. To address this issue, many researchers created platforms focused on physical robustness, low-cost, and modularity \cite{Katz2019,Oncilla}. 
\begin{figure}
\centering
    \includegraphics[width=3.0in]{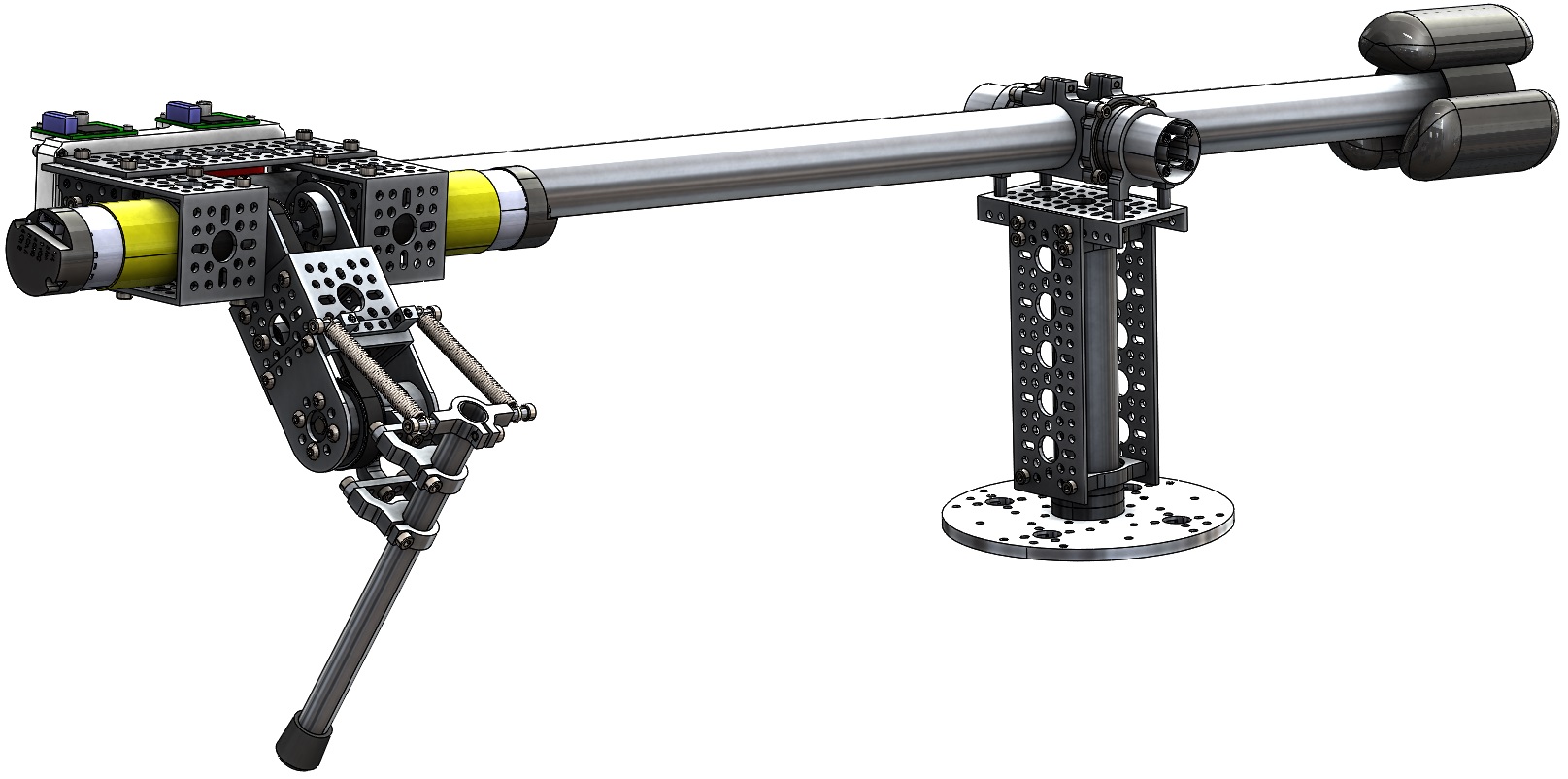}
    \caption{The hopping robot HOPPY for hands-on education in dynamic control and legged locomotion. Available at \url{https://github.com/robodesignlab/hoppy-project}.}
    \label{fig_RobotKit}
\end{figure}

So why create another open-source robot for education? Although remarkable platforms such as the \textit{Open Dynamics Solo} (\$4800) \cite{OpenDynamic} and the \textit{Stanford Doggo} (\$3000) \cite{Doggo} have a significantly lower cost than most legged robots used for research, their use in \textit{large-scale} education is still impractical. Firstly, the base cost of acquiring tens of robot units for a typical robotics class would be formidable. In addition, fabrication equipment is needed, such as CNC machining and 3D printing, and impractical for most courses. Moreover, these robots demand base-level knowledge of mechanical fabrication and electronic prototyping, in addition to low-level programming skills for communication, controls, and simulation. These requirements prohibit the use of existing open-source robot platforms in large cross-disciplinary Engineering courses, especially at the undergraduate level. Feasible educational kits must be student-friendly, mechanically robust, and easy to fabricate, maintain, and program \cite{Hapkit}. Other commercial kits such as the \textit{Robotis Biolod} (\$1,200), and \textit{SoftBank's NAO} (\$14,000) have easy-to-use hardware and software, but are driven by highly geared servo motors that limit their ability to perform dynamic motions like hopping \cite{WensingProprio}. On the other hand, toy-grade products such as \textit{LEGO Mindstorms} (\$350) with plastic parts, do not have the mechanical robustness to sustain continuous impacts inherent to dynamic legged locomotion. To fill the gap in large-scale robotic education, this paper introduces HOPPY (Fig. \ref{fig_RobotKit}), a robust and low-cost kit designed for studying robotics through dynamic legged locomotion. The kit costs under \$500, does not require any fabrication skills, and can be built \textit{exclusively} with off-the-shelf components, allowing the students, not the instructor, to assemble the kit and making it a viable option for online and remote education. In addition to the hardware, the kit includes a physics simulator based on MATLAB, a software adopted by most Universities. The kit covers many of the topics of fundamental robotics courses such as kinematics, dynamics, position and force control, trajectory planning, physics simulation, and more. Although sophisticated controllers can be used to push robot performance, students can achieve stable dynamic hopping with simple heuristics \cite{RaibertBook}, as illustrated by the simple controller we present. As a tool for nurturing active-learning in robotics, the kit was implemented as a hands-on and semester-long project for the Spring 2020 \textit{Robot Dynamics and Control} (ME446) course at the University of Illinois at Urbana-Champaign (UIUC) (\url{http://youtu.be/6O6czr0YyC8}). This hands-on approach to education is becoming increasingly popular due to its proven effect on enabling enduring, deeper, and more significant learning \cite{EduEng,Hapkit2019}.

The contributions of this manuscript include open-sourcing the robot mechanical, kinematic, and dynamic models, a baseline hopping controller, its MATLAB-based simulator with instructions, the bill of materials (BoM), and assembly instructions videos. In addition, we list the basic robotics-related topics we covered in the course and how they relate to HOPPY. We suggest hardware modifications for other educational applications and possible improvements if the budget permits. Finally, we discuss our experience implementing the kit in the ME446 course at UIUC and address future work. This manuscript and the kit are intended for educators interested in implementing \textit{hands-on} robotics activities, and for students, researchers, and hobbyists interested in \textit{experimentally} learning about dynamic robots and legged locomotion. 

\section{Kit design and components}
HOPPY is designed exclusively with off-the-shelf components to lower its cost, enable modularity, customization, and facilitate maintenance. In addition, it is lightweight (total weight about $3.8kg$), portable, and mechanically robust. All components withstanding heavy load are designed to be metal parts available from \url{https://gobilda.com/}. We intentionally avoid the use of plastic 3D printed parts for their short working life under impact loads. Because of the kit's durability, only the initial cost for setting up the course is required and the equipment can be reutilized in future semesters with minimal maintenance costs. The kit available at \url{https://github.com/robodesignlab/hoppy-project} includes: 
\begin{itemize}
    \item Complete list of components and quantities (BoM)
    \item CAD model files in SolidWorks2019 and .STEP
    \item Video instructions for mechanical assembling (\url{https://youtu.be/CDxhdjob2C8})
    \item List of parameters for rigid-body dynamics (link dimensions, mass, center of mass location, and inertia tensor)
    \item Kinematic and dynamic models
    \item The baseline controller described in Section \ref{Sec_Control}
    \item The MATLAB code for dynamic simulation with instructions (version 2016a or later)
    \item Electrical wiring diagram and basic code for the microcontroller ($\mu$C).
\end{itemize}

\subsection{Mechanical System}
\begin{figure}
\centering
    \includegraphics[width=3.4in]{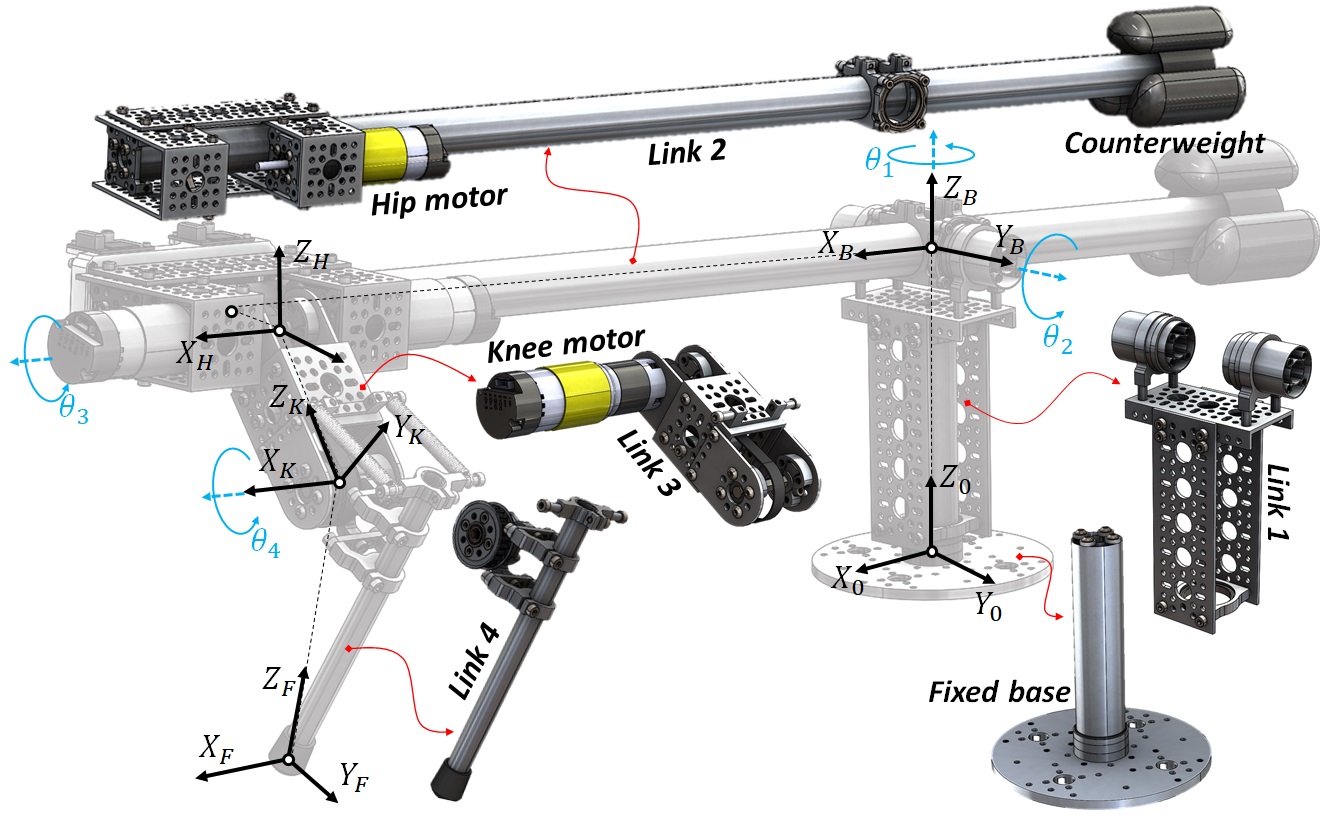}
    \caption{Kinematic transformations and the five rigid-bodies which compose HOPPY. The first two joints ($\theta_1$ and $\theta_2$) are passive while joints three and four are driven by electric motors.}
    \label{fig_Kin}
\end{figure}

Fig. \ref{fig_RobotKit} shows that the mechanical system of the HOPPY kit consists of a robot leg attached to a gantry, similar to the experimental set up for legged robot research \cite{ding2017design} \cite{ding2018single}. The two joints of the gantry ($\theta_{1,2}$) are passive, allowing the robot to translate in a circle and jump vertically. The hip motor on joint 3 ($\theta_{3}$) drives link 3, while the knee motor on joint 4 ($\theta_{4}$) drives link 4. As shown in Fig. \ref{fig_Kin}, both motors are placed near the hip joint to reduce the moving inertia of the leg and enable fast swing motions. The second actuator is mounted coaxially with the hip joint and drives the knee joint through a timing belt. The motors have a minimal gearing ratio in order to be backdrivable and mitigate impacts with the ground \cite{WensingProprio}. To reduce the torque requirements for hopping, springs are added in parallel with the knee, and a counterweight is attached at the opposite end of the gantry. We employ cheap $2.3kg$ ($5lb$) weights typically utilized for gymnastics. The addition of an excessively heavy counterweight reduces the achievable frictional forces between the foot and the ground, and the robot is more likely to slip. However, this low-actuation design strategy provides inherent safety for human-robot-interaction, and elongates the working life of the system.

\subsection{Electrical System}
\begin{figure}
\centering
    \includegraphics[width=2.8in]{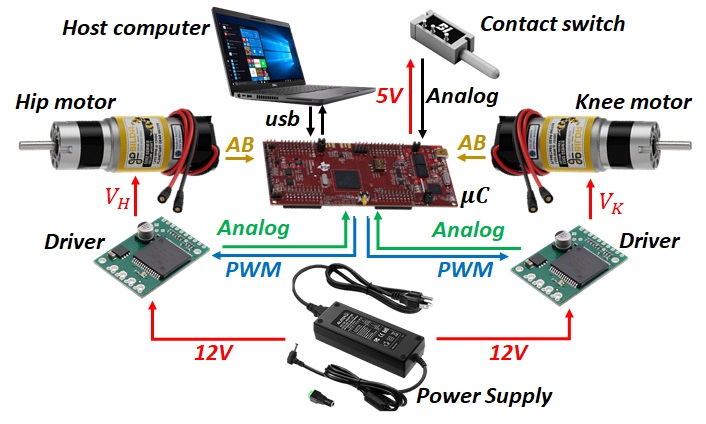}
    \caption{Electrical components of the kit and communication diagram.}
    \label{fig_Electrical}
\end{figure}

The diagram of electrical connections and communications protocols are depicted in Fig. \ref{fig_Electrical}. The user computer communicates with the embedded $\mu$C (\textit{Texas Instruments LAUNCHXL-F28379D}) via a USB port. The $\mu$C commands the desired voltage to the motors (\textit{goBilda 5202-0002-0027} for the hip and \textit{0019} for the knee) via Pulse Width Modulation (PWM) to the drivers (\textit{Pololu VNH5019}) and receives analog signals proportional to the motor current and the pulse counts from the incremental encoders. The motor drivers are powered by a 12V power supply (\textit{ALITOVE 12Volt 10 Amp 120W}). A linear potentiometer with spring return (\textit{TT Electronics 404R1KL1.0)} attached in parallel with the foot works as a foot contact switch.   

\subsection{Modularity and customization}
The kit is designed to facilitate the adjustment of physical parameters and the replacement of components. For instance, \textit{goBilda} provides motor options with a wide range of gear ratios from $3.7:1$ to $188:1$ to be easily adapted to the HOPPY system according to the needs. The length of the lower leg is continually tunable between $60mm$ and $260mm$, and knee springs can be added or removed as deemed necessary. The gantry length is also continually adjustable between $0.1m$ and $1.1m$ and its height has multiple discrete options between $0.2m$ and $1.1m$. The dynamic effect of the counterweight can be adjusted by shifting its position further away from the gantry joint or adding more mass.

\section{Education with Dynamic Legged Robots}
This section describes how the HOPPY kit supports the curriculum of typical fundamental robotics courses. As shown in Fig. \ref{fig_Kin}, the HOPPY platform consists of a two degrees-of-freedom (DoF) active leg attached to a supporting gantry with two passive DoFs, forming a serial chain. To make HOPPY successfully hop, students employ concepts of classical Robotics textbooks \cite{Spong,craig2009introduction,siciliano2010robotics,ModernRobotics} including:
\begin{enumerate}
    \item Coordinate frames and transformations
    \item Forward, inverse, and differential kinematics
    \item Forward, inverse, and contact dynamics
    \item Actuation, gearing, and friction
    \item Trajectory planning
    \item Position and force control
\end{enumerate}
HOPPY can be integrated into a robotics course curriculum as a semester-long project. During the first half of the course, students derive the mathematical models and explore the simulator while assembling their kits. The second half of the course is dedicated to experimentation and exploration. Section \ref{sec_Experiment} describes our approach using this format at UIUC.

\subsection{Kinematics, Dynamics, and Actuation}
The complete kinematic transformations and dynamic equations of HOPPY can be found in the separate document \textit{HOPPY\_Technical\_Guideline.pdf} in the repository, and here we summarize the key concepts. The coordinate frames of HOPPY are presented in Fig. \ref{fig_Kin}, where the origin coordinate frame at the base of the rotating gantry is fixed to the ground. Joints $\theta_{1,2}$ are passive while joints $\theta_{3,4}$ are driven by electric brushed motors. These angles define the robot joint configuration vector $q=[\theta_1 \quad \theta_2 \quad \theta_3 \quad \theta_4]^\top$ and the coordinate transformations between link-fixed frames. The rigid-body properties of each link are provided to the students to derive the standard dynamic equations of motion (EoM) from energy principles using the system's Lagragian. The derived \textit{manipulator equations} \cite{Spong} is:
\begin{equation}\label{eq_EOM}
   M(q)\ddot{q} + C(q,\dot{q})\dot{q} + G(q) = B_e u + J_c^\top F_{GRF},
\end{equation}
where $\dot{q}$ and $\ddot{q}$ are the joint velocity and acceleration; $M(q)$ represents the symmetric and positive definite inertia matrix as a function of the robot joint configuration $q$; $C(q,\dot{q})$ is the Coriolis and centrifugal forces matrix; $G(q)$ is the vector of torques due to gravity; and $u$ is the vector of input torques, including the torque exerted by the knee spring. The spring attached in parallel with the knee alleviates the joint's torque requirements for hopping. Matrix $B_e$ maps the two-dimensional hip and knee torque vector $u$ to the full four-dimensional system dynamics. The robot is underactuated because the system has four degrees-of-freedom but only two control inputs \cite{Underactuated}. Finally, the matrix $J_c$ is the foot contact Jacobian and $F_{GRF}$ represents the ground reaction forces between the foot and the ground. 

Since hopping is a physically demanding task, the actuator limits play a significant role in the stability and performance of the robot. Students could explore how the force and torque requirements of the gait push the velocity and torque boundaries of electric rotary actuators. In addition, the friction in the gearbox is added to the EoM as well as the reflected inertia of the rotors. These governing equations are implemented in the MATLAB-based physics simulator provided with the kit. The simulator serves as a sandbox where students gain familiarity with the fundamentals of the robot kinematics and dynamics before experimenting with the real hardware.

\subsection{Planning and Control}
\label{Sec_Control}
Despite the challenges in legged locomotion control, HOPPY is designed such that a simple heuristic could be employed to achieve stable hopping. The robot alternates between the stance phase and the aerial phase and dedicated controllers are be promptly switched. We employ a simple strategy where the robot applies a prescribed force profile during the stance phase and holds a desired leg configuration during the aerial phase. Because the gantry is designed to be sufficiently long, HOPPY is assumed to perform a motion that is constrained to the $Y_HZ_H$ plane \cite{GrizzleBook}. Further discussion and insights on intuitive hopping control are provided in \cite{Raibert_RunningFast, RaibertBook}. 

\subsubsection{Stance Phase}
During the stance phase, the robot leg applies a pre-defined force profile against the ground, which results in a net impulse on the robot. The force profile is generated using simple functions, such as polynomials for fast online computation. Here we utilize Bézier polynomials to create a force profile for the vertical and horizontal components of the ground contact force within a projected stance duration of $0.15s$ as shown in Fig. \ref{fig_FSM}. The peak value of the force profile is modulated to regulate jumping height and forward velocity.
The desired contact force during stance $F_d$ is mapped to the required joint torques $u_{st}$ using the foot contact Jacobian:
\begin{equation}
    u_{st} = J_c^\top F_{d}.
    \label{eq_TorqueMap}
\end{equation}
Equation \eqref{eq_TorqueMap} assumes that the inertial effects of the lightweight leg in respect to the body are negligible when compared to the contact forces that propel the robot \cite{RaibertBook}. However, more advanced control strategies can be used to incorporate the inertial properties of the leg.

\subsubsection{Aerial Phase}
When HOPPY is in the aerial phase, the leg is held at a desired configuration in respect to the hip in preparation for touchdown. To achieve this, a simple task-space Proportional-Derivative (PD) controller is implemented:
\begin{equation}
    u_{ar} = J^\top_c\left[K_{p}(p_{ref}-p_{f}) - K_{d}\dot{p}_{f}\right],
    \label{eq_aerial}
\end{equation}
where $u_{ar}$ is the vector of joint torques in aerial phase;  $K_{p}$ and $K_d$ are diagonal and positive definite gain matrices, $p_{ref}$ is the desired foot position with respect to the hip at touchdown; $p_{f}$ and $\dot{p}_{f}$ are the current foot position and velocity. More sophisticated controllers such as workspace inverse dynamics control could be implemented to improve the tracking performance.

\subsubsection{Touch-down, Impact, and Lift-off}\label{sec:events}
When the touch-down event is detected, the robot transitions from the aerial phase controller to the stance phase controller. The default contact sensor with the kit is a spring-loaded linear potentiometer (see Fig. \ref{fig_Electrical}) attached to the foot. However, other more capable options exist \cite{Chuah_FootSensor}. Alternatively, the robot can detect contact with the ground by measuring the deviation of the foot from the desired position after impact \cite{Bledt_Contact}. The impact at touch-down instantaneously dissipates part of the total mechanical energy of the system. The MATLAB simulator incorporates a hard contact model \cite{GrizzleBook} for the impact mechanics and assume an inelastic collision that instantaneously brings the foot to a complete stop. The robot controller in the real hardware is oblivious to impact forces due to the fast propagation of shock loads. Instead, large impact loads are avoided by minimizing the reflected inertia of the foot during touchdown \cite{ModernRobotics}. The lift-off event is declared when the contact sensor returns to its original unloaded state.


\subsubsection{Finite State Machine (FSM)}
\begin{figure}
\centering
\resizebox{1\linewidth}{!}{\includegraphics{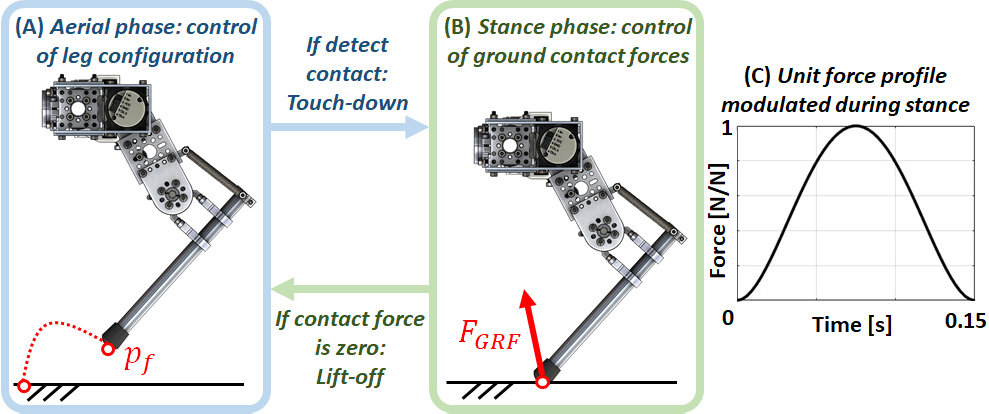}}
    \caption{The finite state machine for the hopping control framework. The robot alternates between (A) aerial and (B) stance phases with transition events of lift-off and touch-down. (C) Prescribed force profile for the horizontal ($Y_H$) and vertical ($Z_H$) components of $F_{GRF}$ during a projected stance of $0.15s$. The peak horizontal and vertical forces are modulated to control locomotion speed and hopping height, respectively.}
    \label{fig_FSM}
\end{figure}

The hopping control framework is summarized as an FSM shown in Fig. \ref{fig_FSM}. This FSM consists of two states, namely, the stance phase and the aerial phase. The transition events, lift-off, and touchdown are described in the previous Section \ref{sec:events}.

\subsection{Hands-on Robotics Labs}
Since HOPPY is tailored for active learning via experimentation, students are exposed to challenges inherent to real robotic systems. Some of the topics are: 

\subsubsection{Embedded and Discrete control} The robot controller runs on an onboard microcontroller ($\mu$C) which interfaces with peripherals and compute inputs at discrete time instances. The deterministic execution of these events is fundamental for the implementation of discrete control policies. To effectively control dynamic motions, the targeted control rate is in the order of $1~kHz$.

\subsubsection{Communication Protocols} 
The $\mu$C receives a digital signal from the incremental encoders, an analog signal proportional to the motor current, and a binary signal from the foot contact switch sensor. The $\mu$C regulates the motor voltage via PWM. 

\subsubsection{Signal Processing} The encoder has a coarse resolution of 28 counts per revolution (CPR) and the gearbox has a noticeable backlash. To avoid noise amplification, motor velocity is estimated using a filtered derivative $\frac{\dot{\theta}(s)}{\theta(s)} = \frac{\lambda s}{s + \lambda}$ converted to discrete-time with a period of $T=1ms$, where $s$ is the Laplace variable for frequency, and $\lambda$ is a tunable constant (usually $\lambda\approx10$). Fast sampling rates are essential to avoid delayed velocity estimation. The analog signal from the motor current estimation is also noisy and requires filtering if used for feedback control.

\subsubsection{Control Input Saturation} The achievable joint torques and speeds are limited by the available voltage supply ($V_{max}=12V$) and the maximum current that the driver can provide ($I_{max}=30A$) \cite{ModernRobotics}.

\subsection{Advanced Topics}
Compared with manipulators commonly used as examples in conventional robotics classes, the behavior of HOPPY is richer and the control more challenging because it \textit{must be dynamic}. In addition, HOPPY continuously makes and breaks contact with the ground, which characterizes it as a system with \textit{hybrid dynamics} \cite{GrizzleBook}. The hybrid nature of legged locomotion provides the students with hands-on experience not easily obtained when working with manipulators. Moreover, the heuristic-based hopping control framework presented here is the basis on which more advanced topics such as model-based control, trajectory optimization, state estimation, and machine learning can studied. Advanced approaches tailored for hybrid dynamical systems, such as limit cycles, Poincare Return Maps \cite{Underactuated}, and Hybrid Zero Dynamics approaches \cite{Grizzle_HZD} can also be explored. Finally, thanks to the modularity of HOPPY, the effect of mechanical design, for instance by varying link dimensions and the gearing ratio of the actuators, can be evaluated with the real robot \cite{Wensing_CoDesign}. Finally, additional modules such as cameras and LIDARs can also be used. 


\section{Limitations, Improvements, and Safety}
\label{sec_Modifications}
Despite the advantages of HOPPY, the inexpensive components employed also have limitations. For instance, voltage control of the motor is not ideal as it hinders the precise and rapid regulation of ground reaction forces. The coarse resolution of the encoders and the backlash in the gearbox also impair position tracking performance. The hobby-grade brushed motors have limited torque density (peak torque divided by mass) \cite{Seok2013}. Thus, the robot \textit{will not} be able to hop without the aid of the springs and the counterweight. In addition, the simulator assumes simple dynamic models for impact and contact and neglects friction, structural compliance, vibrations, and foot slip. Due to these limitations, it is \textit{unlikely} that the physical robot will behave exactly as the simulator predicts. Nevertheless, the simulator is a valuable tool to obtain insights into the behavior of the real robot, and it is useful for preliminary tuning of controllers before implementation in the hardware. Therefore, the robot simulator is a fundamental tool for teaching.

We note that the kit described in this paper is the \textit{bare minimum} required to enable robot functionality. However, if the budget allows, many components of HOPPY can be upgraded. A first priority improvement would be (i) enabling unrestricted rotations around the gantry using an onboard battery to power the robot and a USB slip ring on the first joint for communication between the $\mu$C and the host's computer. The battery can also be used as the gantry counterweight. In addition, (ii) employing encoders with a finer resolution of around 12bits (4096 counts per revolution) would improve position control and joint velocity estimation. (iii) Implementation of motor drivers which can perform high-frequency current control, such as the \textit{Advanced Motion Control AZBDC12A8} would enable precise torque control and improved force tracking during stance. An even better solution would be (iv) the utilization of brushless motors similar to those employed in \cite{OpenDynamic,Katz2019}. These have superior torque density and would enable hopping without the aid of springs or counterweights. (v) Additional sensors (encoders) could be added to the gantry joints ($\theta_{1,2}$) or an Inertial Measurement Unit (IMU) can be mounted on the robot for state estimation \cite{CassieStateEstimation}. And finally, (vi) data logging can be done using external loggers such as the \textit{SparkFun OpenLog}.

When experimenting with HOPPY (or any dynamic robot), safety guidelines must be followed. Students should wear safety goggles and clear the robot's path. It is advised to \textit{constantly} check the temperature of the motor's armature. If they are too hot to be touched, the experiments should be interrupted until they cool down. Aggressive operations (high torque and speeds) within negative work regimes should be done carefully because electric motors regenerate part of the input energy back into the power supply when backdriven \cite{Seok2013}. Therefore, utilizing a battery is more appropriated if considerable negative work is expected. For instance, a conventional $12V$ car battery can be used both as a power supply and a mechanical counterweight. Users should be mindful of the power and USB cables to not be pinched by the joints or excessively twisted during experiments. Pre-winding the cables around the gantry in the opposite direction of motion before experiments may help mitigate this issue. The $\mu$C can also be programmed to run standalone. Finally, applying thread locker to fasteners is recommended during assembly and we advise to often check for loose components due to the constant impacts and vibrations.


\section{Experiments and Discussion}
\label{sec_Experiment}
\begin{figure*}
\centering
    \includegraphics[width=7in]{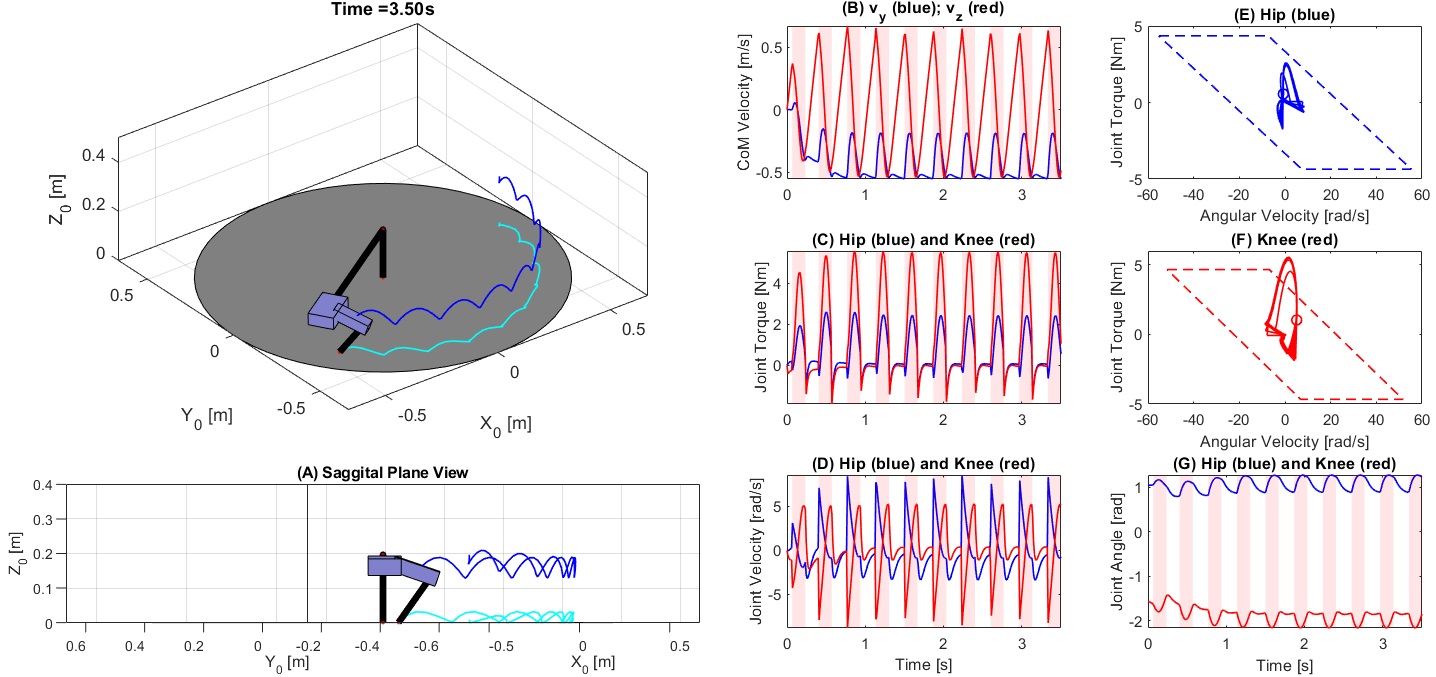}
    \caption{Snapshot of the simulator GUI. It shows the robot animation in isometric and Sagittal plane views along with the trajectories of the hip joint in blue and the foot in cyan. Plot (B) shows the horizontal and vertical tangential velocities of the hip base around the gantry. Areas shaded in red represent stance phases. Hip and knee joint torques are shown in (D) and their angular velocities in (D). Plots (E) and (F) display the polygons that define the limits of the hip and knee actuators along with their torque versus velocity trajectory. Feasible trajectories must be always contained inside the polygon. Finally, plot (G) shows the hip an knee joint angles over time.}
    \label{fig_SimScreenshot}
\end{figure*}
A snapshot of the output of the MATLAB-based simulator is shown in Fig. \ref{fig_SimScreenshot}. It displays a 3D animation of the robot during hopping with a trace of the hip joint and foot spatial trajectories. The view of the robot's planar motion is shown in (A). The visualizer displays the hopping velocity in (B), required torques in (C), the joint velocities in (D), the joint angles in (G), and the working regions of the hip (E) and knee (F) actuators. Areas shaded in red represent stance phases. More information about the simulator is provided in instructions \textit{Code\_Instructions.pdf} in the repository.

\begin{figure*}
\centering
    \includegraphics[width=7in]{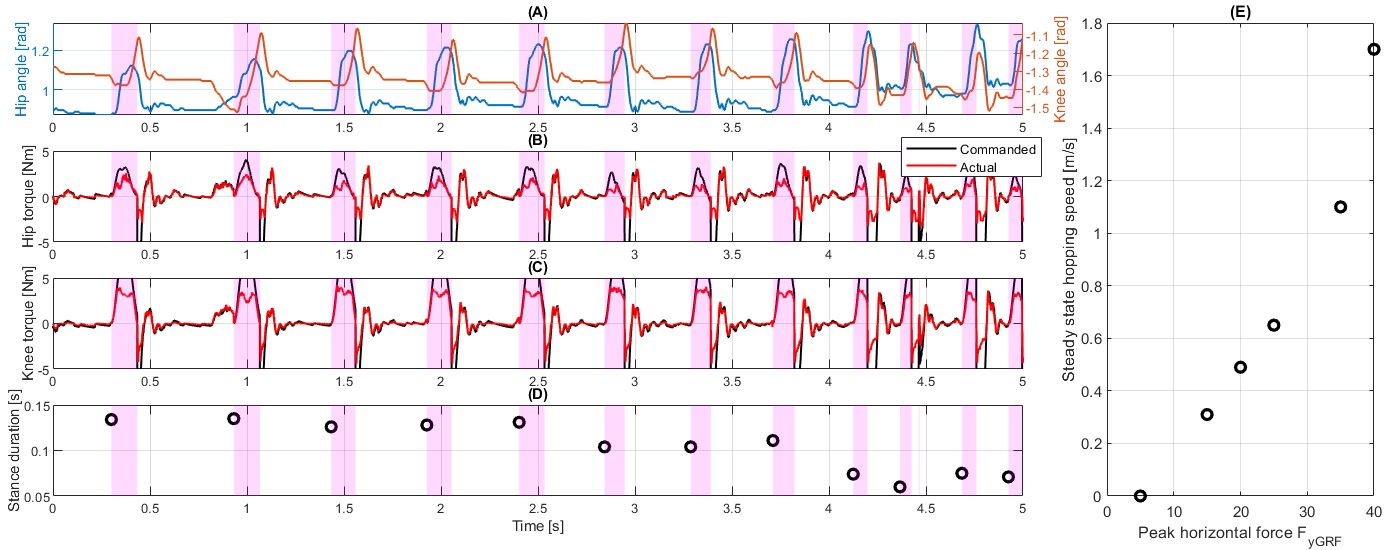}
    \caption{Experimental data with HOPPY speeding up from around $0.65m/s$ to $1.2m/s$. Shaded magenta areas represent stance periods. (A) Hip ($\theta_3$) joint angles in the left vertical axis and knee ($\theta_4$) angle in right axis. (B) Hip joint commanded torque (black) and actual torque (red) estimated from the measured current. (C) Knee joint torque. (D) Stance phase duration (duration of following shaded areas). And (E) steady-state hopping speed as a function of the commanded peak horizontal contact force $F_{yGFP}$.}
    \label{fig_Experiment_Data}
\end{figure*}

For the experiments in companion video (\url{https://youtu.be/AvJIx4CQarM}), we employed the controller described in section \ref{Sec_Control} using aerial phase gains of $K_p = 500[1_{2\times2}]$ and $K_s = 50[1_{2\times2}]$, a prescribed stance duration of $T_s=0.15s$, and a prescribed peak vertical force of $80N$. Where $1_{2\times2}$ is a $2\times2$ identity matrix. The peak horizontal force was modulated to control the locomotion speed. Fig. \ref{fig_Experiment_Data}-(A) to (D) shows the data of the robot accelerating from about $0.65m/s$ to around $1.2m/s$ by increasing the horizontal force peak. The joint torques are estimated by multiplying the measured motor current by the actuator's torque constant. The saturation occurs when the requested input voltage is above the supplied $12V$. Fig. \ref{fig_Experiment_Data}-(D) shows the stance duration, measured from the contact switch, which decreases according to the locomotion speed. Notice that at fast speeds, the robot touches the ground for less than $0.1s$. Finally, \ref{fig_Experiment_Data}-(E) shows the achieved steady-state hopping velocities according to the commanded peak horizontal force. However, we note that the locomotion speed is likely overestimated because it was computed using the relative velocity between the foot and the hip during stance, assuming that the foot does not slip during contact.   

\begin{figure*}
\centering
    \includegraphics[width=7in]{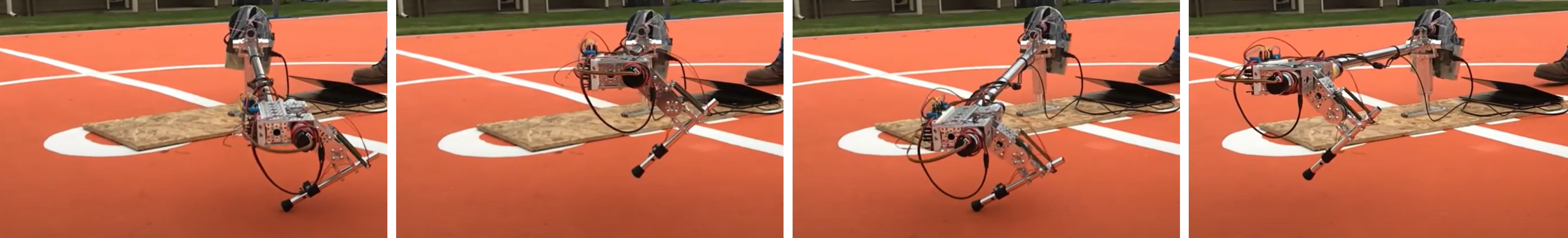}
    \caption{Snapshots of the hopping experiment with HOPPY by one of the student teams (\url{https://youtu.be/6O6czr0YyC8}).}
    \label{fig_Experiment}
\end{figure*}

HOPPY was used in a semester-long project in the \textit{ME446 Robot Dynamics and Control} course at UIUC. The course is attended by senior undergraduate and first-year graduate students from the Departments of Mechanical Science and Engineering, Electrical and Computer Engineering, and Industrial \& Enterprise Systems Engineering. Ten cross-departmental teams of four to five students each received their own leg (hip base of link 2 plus links 3 and 4), and all teams shared six gantries (fixed base plus links 1 and 2). During the first half of the course, students derived the kinematic and dynamic model of the robot, developed different controllers, and created a simple version of the robot simulator in MATLAB. They assembled their kits half-way through the semester and performed experiments with the robot in the second half. After developing their own dynamic simulator, students were provided with our code for comparison. The results from experiments with the real robot from one of the teams are shown in the snapshots of Fig. \ref{fig_Experiment}. The team developed a custom foot contact sensor to detect the stance phase. The robot was able to regulate hopping speed, overcome small obstacles, and mitigate disturbances from gentle kicks. With this hands-on kit, we aim for students to develop a better understanding of the applications of complementary tools (mathematical model, computer simulation, and physical prototype) used to create real dynamic robots.


\section{Conclusion and future work}
This paper introduced HOPPY, an open-source, low-cost, robust, and modular kit for robotics education with dynamic legged robots. The goal of the kit is to lower the barrier for studying dynamic robot behaviors and legged locomotion with real systems. The control of dynamic motions present unique challenges to the robot software and hardware, and these are often overlooked in conventional robotics courses, even those with hands-on sessions. Here we describe the topics which can be explored using the kit, list its components, discuss preferred practices for implementation, and suggest further improvements. HOPPY was utilized as the topic of a semester-long project for a first-year graduate-level course at UIUC. Students provided positive feedback from the hands-on activities during the course. The instructors will continue to improve the kit and course content for upcoming semesters. To nurture active learning, future activities will include a friendly competition between teams to elect the fastest robot and the most energy-efficient robot. Speed is estimated by timing complete laps around the gantry. Energy efficiency is determined by the Cost of Transport, which is proportional to the ratio between the total electrical power consumed during one hopping cycle and the average translation speed \cite{Seok2013}. All future improvements to the kit will be made open-source.


\bibliographystyle{plain}
\bibliography{References_Ramos}
\end{document}